\documentclass{article}

\usepackage{arxiv}

\usepackage[utf8]{inputenc} 
\usepackage[T1]{fontenc}    
\usepackage{hyperref}       
\usepackage{url}            
\usepackage{booktabs}       
\usepackage{amsfonts}       
\usepackage{nicefrac}       
\usepackage{microtype}      
\usepackage{lipsum}		
\usepackage{graphicx}
\usepackage{natbib}
\usepackage{doi}

\usepackage{multirow}
\usepackage{caption}
\usepackage{subcaption}
\usepackage{amsmath}
\usepackage{amssymb}
\usepackage{pifont}
\newcommand{\cmark}{\ding{51}}%
\newcommand{\xmark}{\ding{55}}%

\title{A Large-scale Evaluation of Pretraining Paradigms for the Detection of Defects in Electroluminescence Solar Cell Images}

\author{ \href{https://orcid.org/0000-0003-2822-7146}{\includegraphics[scale=0.06]{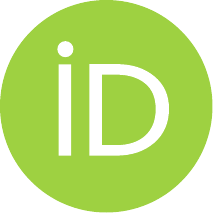}\hspace{1mm}David ~Torpey}\thanks{Alternative email address: 674425@students.wits.ac.za.} \\
	School of Computer Science and Applied Mathematics\\
	University of the Witwatersrand, Johannesburg\\
	South Africa \\
	\texttt{torpey.david93@gmail.com} \\
	\And
	\href{https://orcid.org/0000-0001-6488-1263}{\includegraphics[scale=0.06]{orcid.pdf}\hspace{1mm}Lawrence ~Pratt} \\
	School of Computer Science and Applied Mathematics\\
	University of the Witwatersrand, Johannesburg\\
	South Africa \\
	\texttt{lpratt@csir.co.za} \\
	\And
	\href{https://orcid.org/0000-0003-0783-2072}{\includegraphics[scale=0.06]{orcid.pdf}\hspace{1mm}Richard ~Klein} \\
	School of Computer Science and Applied Mathematics\\
	University of the Witwatersrand, Johannesburg\\
	South Africa \\
	\texttt{kleinric@gmail.com} \\
}

\renewcommand{\shorttitle}{Large-scale EL SCDD Evaluation}

\hypersetup{
pdftitle={A Large-scale Evaluation of Pretraining Paradigms for the Detection of Defects in Electroluminescence Solar Cell Images},
pdfsubject={cs.CV},
pdfauthor={David ~Torpey, Lawrence ~Pratt, Richard ~Klein},
pdfkeywords={Deep learning, Self-supervised learning, Benchmark datasets, Semantic segmentation, Electroluminescence images},
}

\begin{document}
\maketitle

\begin{abstract}
Pretraining has been shown to improve performance in many domains, including semantic segmentation, especially in domains with limited labelled data. In this work, we perform a large-scale evaluation and benchmarking of various pretraining methods for Solar Cell Defect Detection (SCDD) in electroluminescence images, a field with limited labelled datasets. We cover supervised training with semantic segmentation, semi-supervised learning, and two self-supervised techniques. We also experiment with both in-distribution and out-of-distribution (OOD) pretraining and observe how this affects downstream performance. The results suggest that supervised training on a large OOD dataset (COCO), self-supervised pretraining on a large OOD dataset (ImageNet), and semi-supervised pretraining (CCT) all yield statistically equivalent performance for mean Intersection over Union (mIoU). We achieve a new state-of-the-art for SCDD and demonstrate that certain pretraining schemes result in superior performance on underrepresented classes. Additionally, we provide a large-scale unlabelled EL image dataset of $22000$ images, and a $642$-image labelled semantic segmentation EL dataset, for further research in developing self- and semi-supervised training techniques in this domain.
\end{abstract}


\section{Introduction}
As the global economy moves towards renewable energy to power the increasingly electrified economy of the future, the energy landscape is constantly evolving. Although power generation from solar photovoltaic (PV) systems accounted for only 3.6\% of the world's electricity in 2021, it did mark a 22\% increase over 2020 following years of steady growth \citep{iea2022pv}. The basic operating principle of a solar cell, known as the `photovoltaic effect', was discovered by Edmond Becquerel in 1839. The solar cell harnesses energy from the sun in the form of photons to generate a voltage and electric current. Most of the PV systems in the world are still built from the silicon solar cell that lies at the core of the electrical energy generator. The sheer volume of solar modules made from individual solar cells poses a challenge for monitoring their quality and reliability, making solar cell defect detection (SCDD) a critical tool for the present and future.

\begin{figure}[!h]
    \centering
    \includegraphics[width=0.45\textwidth]{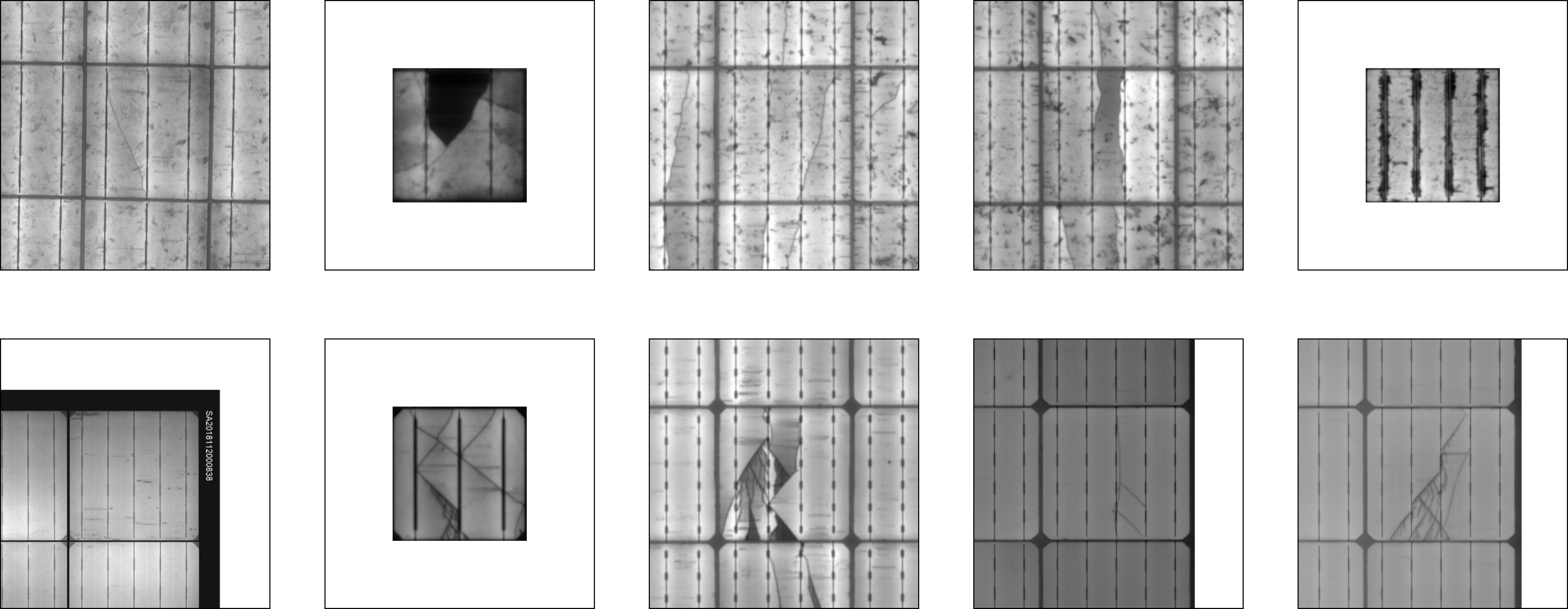}
    \caption{EL images of multi-crystalline (top) and mono-crystalline (bottom) silicon solar cells}
    \label{fig:2x5_el_defects}
\end{figure}

Electroluminescence (EL) images of solar cells are essential for SCDD, as many defects are not visible to the naked eye. An EL image acts like an x-ray for a doctor, enabling the PV expert to identify defects such as cracks and inactive areas in solar cells that can negatively impact performance. Millions of EL images are recorded every year across the globe and analyzed by PV module manufacturers, PV module buyers, PV system operators, and independent test labs. Figure \ref{fig:2x5_el_defects} depicts EL images of multi-crystalline and mono-crystalline silicon solar cells with defects such as cracks, gridline defects, inactive areas, and ribbon corrosion. The U-net and DeepLabV3+ architectures have proven effective at identifying these defects \citep{pratt2021,pratt2023}.

Simultaneously, the field of machine learning has witnessed the development of various models that integrate both labelled and unlabelled data into the training process to enhance performance, particularly in cases where there is limited data for supervised learning regimes. While self-supervised (SelfSL) and semi-supervised learning (SemiSL) methods have been widely used with common benchmark datasets, such as ImageNet \citep{ILSVRC15}, Cityscapes \citep{Cordts2016Cityscapes}, and CIFAR10 \citep{krizhevsky2009learning}, these datasets all comprise natural images of common objects. It is unclear whether pretraining on large-scale, object-centric datasets like ImageNet transfers to downstream semantic segmentation tasks that have a large distribution shift from ImageNet.

To address this gap, we perform a comprehensive analysis of these pretraining paradigms in the context of electroluminescence (EL) solar cell defect detection (SCDD). Given the growing importance of SCDD applications, we present a large-scale benchmark of the existing tools and training recipes that can be used to enhance performance in this domain. Our objectives in this work are to 1) show how the inclusion of unlabelled data in the modelling process can improve SCDD performance, 2) perform a benchmark and analysis of different training paradigms that make use of the unlabelled data against several strong baselines, and 3) contribute two EL datasets to the research community, including a fully-annotated semantic segmentation benchmark dataset and a large-scale unlabelled dataset. It is important to note that the goal of this work is not to develop a new method for SCDD but rather to benchmark common pretraining paradigms, show how the inclusion of unlabelled data can improve the state-of-the-art in the domain, and through this demonstrate the utility of SelfSL still leaves something to be desired when moving to non-ImageNet-like settings such as SCDD.

In particular, we analyse three main paradigms. Firstly, we analyse supervised semantic segmentation. Next, we analyse both in-distribution and OOD self-supervised pretraining using two state-of-the-art methods: SimCLR \citep{simclr} and MoCov2 \citep{mocov2}. Finally, we analyse two semi-supervised semantic segmentation algorithms: CCT \citep{cct} and U2PL \citep{u2pl}. We find three models perform equally well for the SCDD application. Specifically, the supervised training on a large OOD dataset (COCO), pretraining on a larger OOD dataset (ImageNet), and semi-supervised pretraining (CCT) all yield statistically equivalent mIoUs, on average. The semi-supervised model U2PL performed significantly worse than all other models under evaluation.

Our contributions can be summarised as follows:
\begin{itemize}
    \item The first to make use of unlabelled data for SCDD and we achieve a new state-of-the-art in the domain.
    \item A large-scale analysis and benchmarking of pretraining methods for improving SCDD performance.
    \item We show that supervised training on a large OOD dataset (COCO), self-supervised pretraining on a large OOD dataset (ImageNet), and semi-supervised pretraining (CCT) all yield statistically equivalent performance for mIoU. We give preference to the COCO model due to its simplicity, until such time that SelfSL is better suited to the EL domain.
    \item We show that self-supervised pretraining in the SCDD domain does not yield a statistically significant improvement over supervised training, contrary to recent results for ImageNet-like domains \citep{sslanalysis}. This speaks to a need for further research and tailoring of self-supervised techniques for non-ImageNet-like data.
    \item We publish a large-scale unlabelled solar cell image dataset containing $22000$ images, and a labelled dataset with $642$ images and ground truth masks for SCDD for semantic segmentation (available at \textbf{redacted for blind review}).
\end{itemize}

The rest of the paper is structured as follows. Section \ref{sec:related_work} introduces previous work in the area of SCDD, and relevant work in SelfSL and SemiSL. Section \ref{sec:methodology} introduces the approach to performing the experiments and analyses of the different pretraining and fine-tuning paradigms. Section \ref{sec:experiments} contains the results for all experiments and accompanying analyses. Finally, we provide concluding remarks in Section \ref{sec:conclusion}, including potential avenues for future work in this area.

\section{Related Work}
\label{sec:related_work}
\subsection{SCDD}
Many researchers have focused on automated methods for SCDD over the years. However, multi-class semantic segmentation models have received relatively little attention to date, and that is likely due to a lack of relevant data. There are dozens of publicly available datasets for training models to detect natural and man-made objects. The ImageNet \citep{ILSVRC15} dataset consists of 1000 object classes including animals, food, and man-made objects found in everyday modern life. The original COCO \citep{lin2014microsoft} dataset consists of 91 objects in 12 categories also common in modern life. The Cityscapes \citep{Cordts2016Cityscapes} dataset consists of 30 object classes commonly found in modern urban life including roads, buildings, vehicles, and people. The Google Open Images V7 dataset contains approximately 9 million images spanning over 20,000 categories. The CIFAR10 \citep{krizhevsky2009learning} dataset consists of ten classes including animals, aeroplanes, ships, and trucks. The Google Open Images database does include seven categories containing the word 'solar' and over 5,000 corresponding images. The images range from small PV panels for battery charging to large, ground-mounted PV systems. The `solar vehicle' category contains many interesting and novel applications of solar PV on land and water. ImageNet1000 also includes three categories with the word `solar' in the title. However, none of these datasets includes EL images of solar cells which require specialised equipment to obtain.

There are a limited number of public datasets available with EL images of solar cells. \cite{French_Karimi_Braid_2020} published 1,028 EL images of solar cells for the classification of good, cracked, and corroded cells. In \cite{Buerhop2018,Deitsch2019,Deitsch2021} the authors publish a dataset of 2,624 EL images and corresponding defect probabilities for analysis. \cite{pvELad} assembled a database of 36, 543 EL images from multi-crystalline silicon cells with ground truth bounding boxes in 2021. The March 2022 Kaggle competition based on that dataset attracted only limited participants, but the database is reportedly available for research and academic purposes only. Still, none of these datasets enables a semantic segmentation model for SCDD. In 2022, \cite{fioresi2022automated} published a dataset for SCDD including 11,851 EL images and the annotations for training and an additional 17,000 images for testing. The work focused on ten different defect classes associated with contact, interconnection, and cracks. In 2023, \cite{pratt2023} published a dataset for SCDD including 593 EL image annotations to detect 12 defect classes and 12 feature classes. This dataset has since been extended to contain 642 labelled EL images. 

Much of the published research related to SCDD has focused on object detection and classification. Some authors published work on binary segmentation for cracks and gridline defects \citep{tsai2012defect,spataru2016automatic,chen2019accurate,rahman2020defects}. \cite{fioresi2022automated} presented recent work on multi-classification using semantic segmentation for two classes of defects. In 2023, \cite{pratt2023} published a benchmark dataset and performance metrics using semantic segmentation for SCDD. The benchmark dataset consists of 593 images with semantic segmentation labels for 12 features and 12 defects common in solar cells. This dataset is the most detailed semantic segmentation dataset currently available and the benchmarks represent the current state of the art. These two articles included performance metrics from several fully-supervised deep learning models for detection, classification, and quantification of multiple defects in solar cells. The intersection over union (IoU) and recall for specific defects were key metrics for assessing model performance. Cracks and gridline defects with a relatively high occurrence in the benchmark dataset were detected with high recall (0.86 and 0.85, respectively), as were the larger features like ribbons (0.98) and spacing (0.95). The IoU for these same defects and features was low primarily due to dilation in the predictions which drove precision down. In addition, many of the defects and features such as rings, busbars, splices, and text with only a few samples in the training dataset were not detected at all.

One thing common to all previous works in SCDD is the fact that unlabelled samples were never included as part of the modelling process. In this paper, we aim to remedy this by making use of the unlabelled data in both self-supervised and semi-supervised settings.

\subsection{Self-Supervised Learning}
SelfSL is a class of algorithms in which the supervision signal is derived from the data itself, as opposed to human or manual labelling. This inherently makes these algorithms more easily scaled in terms of data size since it removes the requirement of the human annotation bottleneck of conventional supervised learning. There are two common approaches to SelfSL in the computer vision domain: 1) pretext tasks and 2) instance discrimination (ID), with techniques within the ID umbrella currently representing the state-of-the-art. In this work, we experiment with two ID techniques: SimCLR \citep{simclr} and MoCov2 \citep{mocov2}. SimCLR is a contrastive learning method that uses the InfoNCE \citep{infonce} objective to train a network to discriminate between different views (obtained through random data augmentation) of the same or different images. SimCLR performs best with large batch sizes in order to obtain a sufficiently large amount of negative samples when computing the loss. MoCo \citep{moco} is an algorithm that also minimises InfoNCE, however, it contains a Siamese architecture in which one network (encoder) is updated via backpropagation, while the other (momentum encoder) is updated via an exponential moving average of the encoder weights (which is used to ensure the consistency of the representations of negatives samples drawn from a memory bank). MoCov2 simply improves on MoCo by adding a projection head and additional data augmentation to generate random views.

SelfSL models, particularly new models, are typically evaluated on ImageNet and similar object-centric, natural-image datasets such as CIFAR10, Food101, and VOC2012. Studying the behaviour and impact of SelfSL on very different, non-ImageNet-like datasets - such as our EL data - is less well studied, although there have been evaluations and analyses of SelfSL in various domains such as medical \citep{ssl-for-medical} and video game data \citep{phd-jei-midas}. The findings in these studies show that SSL has mixed benefits in these kinds of non-ImageNet domains, sometimes performing well (linear evaluation on video game data or class imbalanced scenarios on medical data) and performing poorly in other cases (fine-tuning for the video game data and balanced or \emph{severely} imbalanced medical data regimes).

\subsection{Semi-Supervised Learning}
While SelfSL techniques are typically applied as pretraining techniques, semi-supervised learning (SemiSL) techniques focus on incorporating both labelled and unlabelled data into the training process. We focus on two SemiSL methods designed for semantic segmentation.

CCT \citep{cct} is a consistency-based semi-supervised technique for semantic segmentation tasks. A shared encoder and main decoder are trained in a supervised fashion, while consistency with the output of various auxiliary decoders is simultaneously enforced. Consistency is enforced by making the model's predictions invariant to small perturbations in the input. U2PL \citep{u2pl}, on the other hand, approaches semi-supervised semantic segmentation via pseudo-labelling \citep{Pseudolabel} rather than consistency training. Typically, the only pseudo-labels that are used for training are those which the model is confident about. However, U2PL aims to make use of these unreliable pseudo-labels, as they may still prove useful for discrimination in the task at hand. This is done by using so-called \emph{anchor} pixels (or \emph{query} pixels), selecting associated negative and positive samples for each anchor pixel, and computing a contrastive loss. This contrastive loss is part of the overall loss function, which includes additional supervised and unsupervised terms.

\section{Methodology}
\label{sec:methodology}
We consider multiple pretraining and fine-tuning setups. The pretraining setups include 1) pretraining on an OOD unlabelled dataset (i.e. self-supervision on ImageNet), 2) pretraining on an in-distribution unlabelled dataset (i.e. self-supervision and semi-supervision on the unlabelled EL dataset), 3) no pretraining (i.e. random weights), and 4) pretraining on an OOD labelled dataset (i.e. ImageNet or COCO). The fine-tuning setup remains consistent in all cases, which is semantic segmentation on the labelled EL dataset.

Formally, consider an OOD unlabelled set of images $X_{ul}^{\texttt{OOD}}$, an in-distribution unlabelled set of images $X_{ul}$, and a labelled set of images for the downstream task $X = \{(x_i, y_i)\}$, where $x \in X$ is an image and $y \in Y$ is its associated semantic segmentation mask. Consider a function that can estimate the segmentation mask given an image $f: \mathcal{X} \rightarrow \mathcal{Y}$.

The first regime (R0) we consider is the current state-of-the-art model for SCDD \citep{pratt2023}. R0 uses a DeepLabV3+ model with a ResNet34 backbone trained on an EL dataset with initial weights transferred from ImageNet. The next regime (R1) is where $f$ is a randomly-initialised DeepLabV3 model. This is where we simply train the model from scratch without any form of pretraining. These two serve as the baseline models. The next regime (R2) is where $f$ is initialised by supervised pretraining on an OOD semantic segmentation dataset, followed by fine-tuning on $X$.

For the remaining regimes, we take advantage of the fact that $f$ can be decomposed into an encoder and decoder: $f = g \circ h$, where $g$ is the decoder and $h$ is the encoder. First, we consider the case where $h$ is first estimated by pretraining on ImageNet in a supervised fashion (R3). The next regimes we consider are where $h$ is first estimated offline by performing self-supervised pretraining using $X_{ul}^{\texttt{OOD}}$ with SimCLR (R4) and MoCov2 (R5), and using $X_{ul}$ with SimCLR (R6). Lastly, we have the semi-supervised regimes, where $f$ is estimated using both $X_{ul}$ and $X$ simultaneously. In this regime, we consider estimating $f$ using CCT (R7) and using U2PL (R8), Finally, we consider the case where the encoder estimated from R7 is used for fine-tuning a DeepLabV3 model using $X$ (R9). We use CCT and U2PL as these are two, recent state-of-the-art methods tailored specifically for semi-supervised semantic segmentation tasks. Additionally, they provide a simple and easy way to standardise the backbone networks to align with the other regimes under consideration. 

For the training of SimCLR in R6, we perform a hyperparameter sweep over the learning rate and weight decay, where the metric is IoU on the validation subset of $X$. The parameters used in the estimation of $f$ for all other regimes are aligned with those reported in the original works \cite{pratt2023,cct,u2pl}.

While we use various initialisation schemes for the encoder, we use random initialisations for the decoder, $g$, in all models except R2 (COCO), where the pretrained decoder weights from torchvision are used.

It should be noted that in our experiments, $|X_{ul}^{\texttt{OOD}}| \gg |X_{ul}| \gg |X|$, where $|\cdot|$ represents the number of images in the dataset. As $X_{ul}^{\texttt{OOD}}$ we employ the ImageNet and COCO datasets. Additionally, we leverage the extended $642$ EL image CSIR dataset, which includes comprehensive semantic segmentation labels for each image. Furthermore, we constructed an unlabelled EL image dataset of 64,000 images from commercial partners. We release $22000$ of these images publicly for future research focused on the advancement of self- and semi-supervised training techniques in this field. The unlabelled dataset is available at \textbf{repository redacted for blind review}. We use a random 5000-sample subset of $X_{ul}$ during the estimation of $f$ in the SemiSL regimes (R7, R8, and R9). For R6 we use the full set. We found this works better empirically for our EL dataset.

\section{Experiments}
\label{sec:experiments}
\subsection{Experimental Setup}
For R0 (SOTA), we used the same setup as the current state-of-the-art for segmentation-based SCDD \citep{pratt2023}. This is a ResNet34 backbone in a DeepLabV3+ initialised with ImageNet weights. It was trained for 1000 epochs, as per the rest of the experiments. For all other regimes, we use a ResNet50 \citep{resnet} backbone with a DeepLabV3 segmentation model, unless otherwise stated. For R1 (Random) we use random weight initialisation. For R2 (COCO) and R3 (ImageNetSup), we use the pretrained models within torchvision \citep{torchvision} where the models were pretrained on COCO (supervised semantic segmentation) and ImageNet (supervised image classification) respectively. For R4 (ImageNet SimCLR) and R5 (ImageNet MoCov2), we use the models available within the VISSL model zoo \citep{vissl} where the backbone models were pretrained using SimCLR and MoCov2 on ImageNet respectively. For R6 (EL SimCLR), we train a SimCLR model on the entire unlabelled in-distribution dataset $X_{ul}$ for 500 epochs with the LARS \citep{lars} optimiser and a batch size of 32. We use an initial learning rate of 0.09, weight decay of 1.4e-07, and 10 warmup epochs. The learning rate is decayed using cosine decay. The default set of data augmentations is used as per the original SimCLR work \citep{simclr}.

For R7 (CCT), R8 (U2PL), and R9 (CCT+), we train at a resolution of 512x512 with the SGD optimiser with a learning rate of 0.01 (with polynomial decay), weight decay of 0.0001, and momentum of 0.9. However, we only train CCT for 400 epochs and U2PL for 500 epochs, as there was no improvement in mIoU past these points for the respective models.

All DeepLabV3 models were trained with the same hyperparameters as the previous SOTA model \citep{pratt2023}. Additionally, we use custom class weights as per that work. We train all DeepLabV3 models for 1000 epochs.

For all models, we report results on the test set for the best model found in terms of mIoU on a validation dataset. We report two metrics: mIoU (mean intersection-over-union) and a balanced version of IoU, which we term wIoU. This is simply IoU weighted according to the frequency of the different classes. This is done in order to include a metric that lessens the importance of low-cardinality classes, which is important in certain parts of our analyses and appropriate in the SCDD domain when there is a large class imbalance (as we have in our test set).

\subsection{Dataset}
\begin{figure}[!h]
    \centering
    \includegraphics[width=0.45\textwidth]{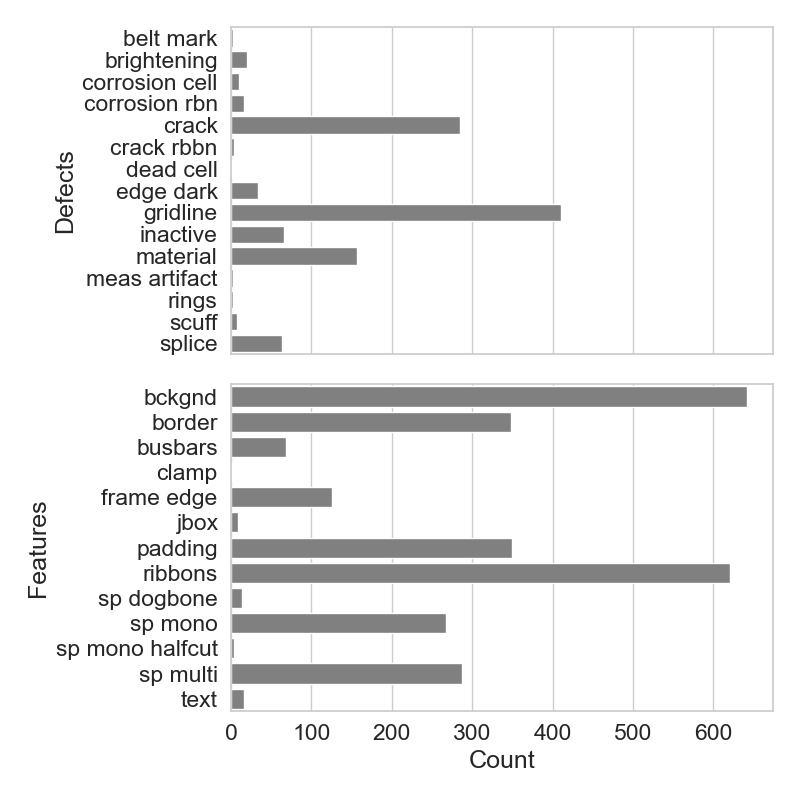}
    \caption{Number of images containing each defect class (top) and feature class (bottom)}
    \label{fig:barchart_counts}
\end{figure}

The experiments were conducted on a dataset developed for SCDD using semantic segmentation. The dataset consists of 642 EL images of solar cells with labelled ground truth masks including 15 defect classes representing potential extrinsic problems with solar cells and 13 feature classes representing intrinsic features of solar cells. Figure \ref{fig:barchart_counts} shows the counts of defects and features represented in the dataset. Some classes are well represented for the purposes of training and statistical analysis of the test results. Some classes are present in only a few images which hinders the training process and analysis.

\begin{figure}[!h]
\centering
\begin{subfigure}[b]{1\columnwidth}
  \centering
  \includegraphics[width=\linewidth]{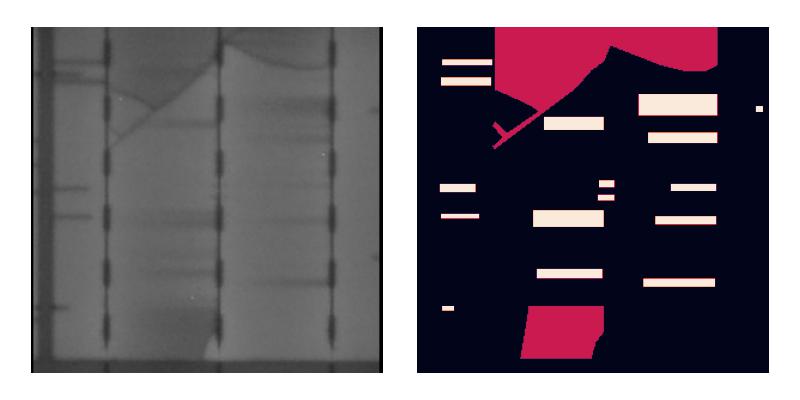}
  \caption{UCF}
  \label{fig:el_gt_ucf}
\end{subfigure}

\begin{subfigure}[b]{1\columnwidth}
  \centering
  \includegraphics[width=\linewidth]{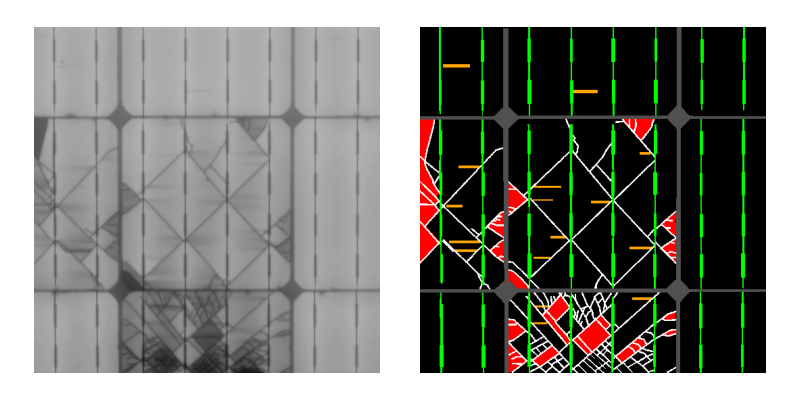}
  \caption{CSIR Dataset}
  \label{fig:el_gt_csir}
\end{subfigure}
\caption{EL and ground truth from UCF \protect\citep{fioresi2022automated} (top) and the CSIR dataset \protect\citep{pratt2023} (bottom) }
\label{fig:el_ground_truth}
\end{figure}

Figure \ref{fig:el_ground_truth} shows examples of EL images and ground truth masks from UCF \citep{fioresi2022automated} and the dataset used in this work. The UCF images exclusively label the extrinsic defects while the ground truth images in this work include defects and the intrinsic features of the solar cells such as cell spacing (grey) and ribbon interconnects (green). By including the intrinsic features in the masks, the potential regions for defect detection are reduced because the feature areas are essentially masked. This may or may not enhance the model's performance, but it provides more context in which the model may learn the defects. The model is trained to see the full image, not just the defects. The ground truth images used in this work also tend to show finer resolution for cracks (thin white lines) and gridline/contact defects. The gridline/contact defects are represented by the thin orange rectangles in the CSIR dataset while they appear as larger rectangular regions in the UCF ground truth. Finer features are more challenging to detect than larger features because they occupy a smaller region in the overall image.

\subsection{Overall SCDD Performance}
We can see in Table \ref{tbl:main_results} that the semi-supervised method CCT (R7) performs best in terms of mIoU, while a COCO-pretrained DeepLabV3 (R2) performs best in terms of wIoU. The top 3 models for mIoU metric, R7, R2, and R4, and interestingly all incorporate supervised pretraining in their regime. The differences among the top three models are not statistically significant when comparing the average mIoU across runs. The same 3 models are also the top performing for wIoU. This suggests that the superiority of self-supervised pretraining over supervised pretraining described in \citep{sslanalysis} does not necessarily generalise to this SCDD domain. We argue that modern SelfSL techniques' objectives are not designed to allow the network's embeddings to effectively capture relevant information for downstream tasks that involve images from a distribution that significantly differs from the one used to train the self-supervised model. However, the supervision signal derived from image classification-based supervised training on ImageNet has been widely shown to perform well on a host of downstream tasks, even where the data is notably different from ImageNet. A typical evaluation of SelfSL versus supervised learning will focus on semantic segmentation datasets that are comprised of common objects and natural scenes \citep{sslanalysis}, which is very different to the EL dataset. R2 and R7 also show a statistically significant increase in mIoU compared to the previous SOTA model R0 ($p < 0.03$). The differences in average mIoU between SOTA and the other models are not statistically significant except for the improved performance of SOTA versus R8 U2PL ($p < 0.002$). 

Notably, the other semi-supervised algorithm, U2PL, performs worst -- even worse than random weights ($p \le 0.01$) for all pairwise comparisons. This phenomenon was noticed in the original U2PL work and occurs when the pseudo-label pixels are not adequately filtered out when using the contrastive loss term. It can lead to worse results than simply training on the supervised data alone. It should also be noted that in regime R9, training the encoder of a trained CCT model in a supervised manner for additional epochs within DeepLabV3 does not improve performance - in fact, both mIoU and wIoU decreased. This suggests that the CCT architecture alone is sufficient to perform semantic segmentation for SCDD.

\begin{table}[!h]
\centering
\caption{Results on the EL dataset for the SCDD semantic segmentation task.}
\label{tbl:main_results}
\begin{tabular}{|l|c|c|c|}
\cline{2-4}
\multicolumn{1}{c|}{} & mIoU & wIoU & OOD \\ \hline
R0 (SOTA) &  0.506    &  0.535    &  \cmark  \\ \hline
R1 (Random) &  0.486    &  0.542    &  -  \\ \hline
R2 (COCO) &  0.544    &   \textbf{0.589}    & \cmark   \\ \hline
R3 (ImageNet Sup) &  0.508    &   0.562    & \cmark   \\ \hline
R4 (ImageNet SimCLR) &   0.534   &   0.578    & \cmark   \\ \hline
R5 (ImageNet MoCov2) &  0.508     &   0.562   & \cmark   \\ \hline
R6 (EL SimCLR) &  0.517     &    0.571   & \xmark   \\ \hline
R7 (CCT) &  \textbf{0.550}     &  0.580      & \xmark    \\ \hline
R8 (U2PL) &  0.448    &  0.515    & \xmark    \\ \hline
R9 (CCT+) &  0.508    &   0.562   & \xmark   \\ \hline
\end{tabular}
\end{table}

\begin{figure}[!h]
    \centering
    \includegraphics[width=\columnwidth]{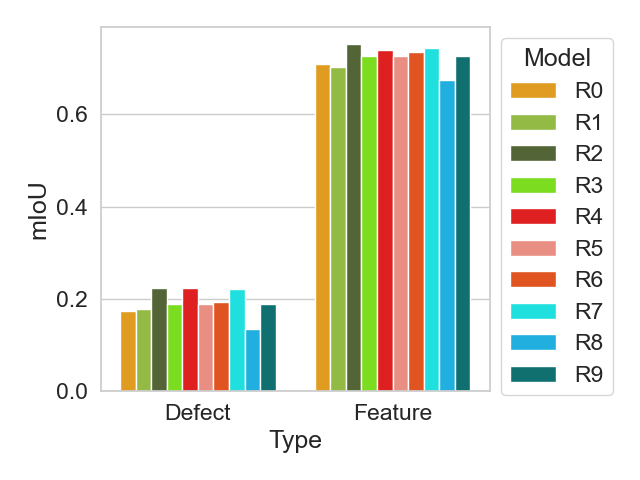}
    \caption{mIoU by class type (extrinsic defect and intrinsic feature).}
    \label{fig:miou_class_type}
\end{figure}

\begin{figure}[!h]
    \centering
    \includegraphics[width=\columnwidth]{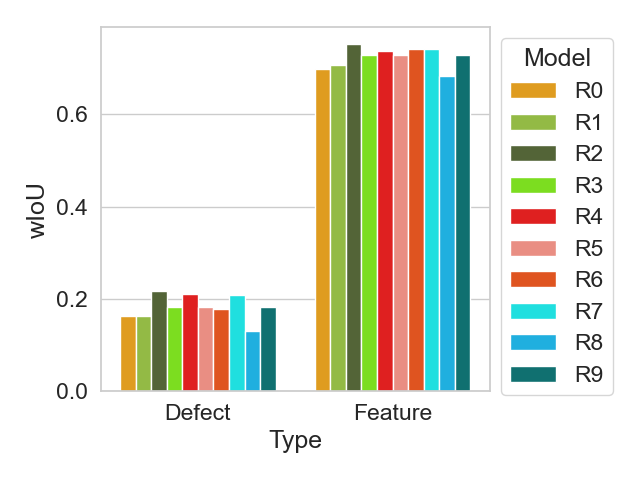}
    \caption{wIoU by class type (extrinsic defect and intrinsic feature).}
    \label{fig:biou_class_type}
\end{figure}

From Figures \ref{fig:miou_class_type} and \ref{fig:biou_class_type} we can see that performance on extrinsic defects is consistently worse than on intrinsic features. The reason for this is likely two-fold. Firstly, defects are less common, and thus this introduces a natural class imbalance. Secondly, and more pertinently, defects are generally smaller in size than features. This makes them harder to detect for conventional semantic segmentation algorithms that are typically optimized for and benchmarked on datasets that contain natural scenes and common objects, such as COCO, Pascal VOC, and CamVID. To quantify this, we compute the average size of a defect versus a feature in the EL dataset: 24\% for a feature and 1.5\% for a defect (measured as a percentage of the image). Therefore, features are roughly 16 times larger than defects.

Nevertheless, it is clear that the trends from Table \ref{tbl:main_results} hold when analysing the Type level. The (semi-)supervised models (R2, R4, and R7) perform best, while U2PL falls behind. It is also interesting to note that there is less variance between model performance for the wIoU metric than for mIoU. This is because there is high variability between the low-cardinality classes and their associated predictions.

\subsection{Defect Analysis}
\begin{figure}[!h]
    \centering
    \includegraphics[width=\columnwidth]{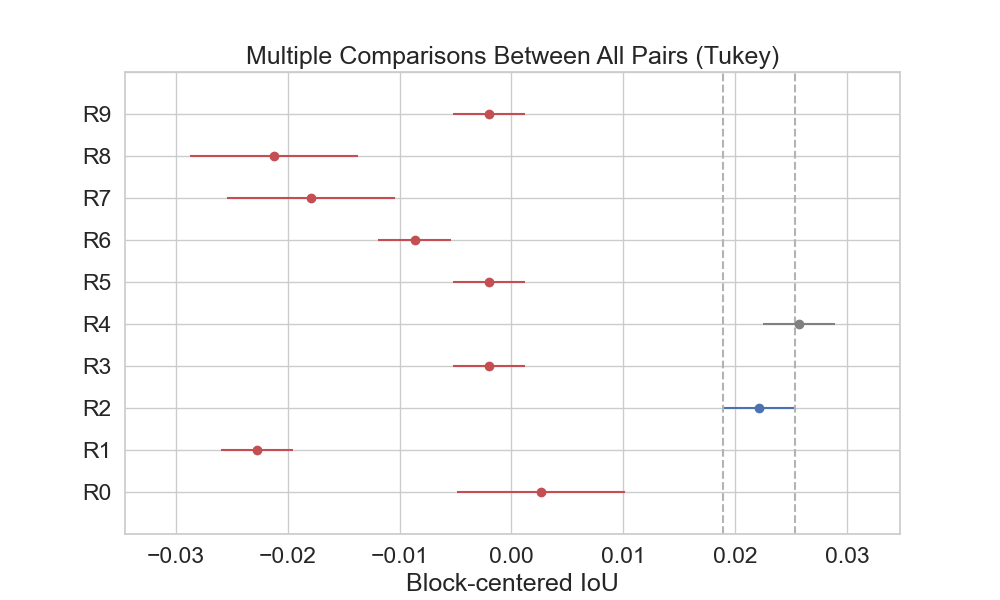}
    \caption{Block-centered IoU average and 95\% confidence intervals for crack detection by regime}
    \label{fig:tukeyHSD_cracks}
\end{figure}

While 15 defects are present in the dataset, crack detection remains paramount in real-world applications. A statistical analysis of the IoU for cracks indicates that R2 and R4 outperform all other models ($p \le 0.0001$ for all pairwise comparisons), including R7 CCT. Figure \ref{fig:tukeyHSD_cracks} shows the mean and 95\% confidence intervals for crack detection based on Tukey's Honest Significant Difference (HSD) as implemented in the `statsmodels' Python library \citep{seabold2010statsmodels}. To improve the signal-to-noise ratio, the image level IoU values were block-centred, i.e. the average value across all runs for each class was subtracted from image-level IoU for each class to block out the high variability from one image to the next for a given class. The block-centered IoU values were then averaged within each regime leading to a sample size of five for each regime with multiple runs. The simultaneous confidence intervals calculated by `statsmodels' are based on the overall variance divided by the number of observations in each group \citep{multicomp}, so the confidence intervals for R0, R7, and R8 are wider than the intervals for other models. R7 and R8 were run only once due to high computational costs.

\begin{figure}[!h]
    \centering
    \includegraphics[width=\columnwidth]{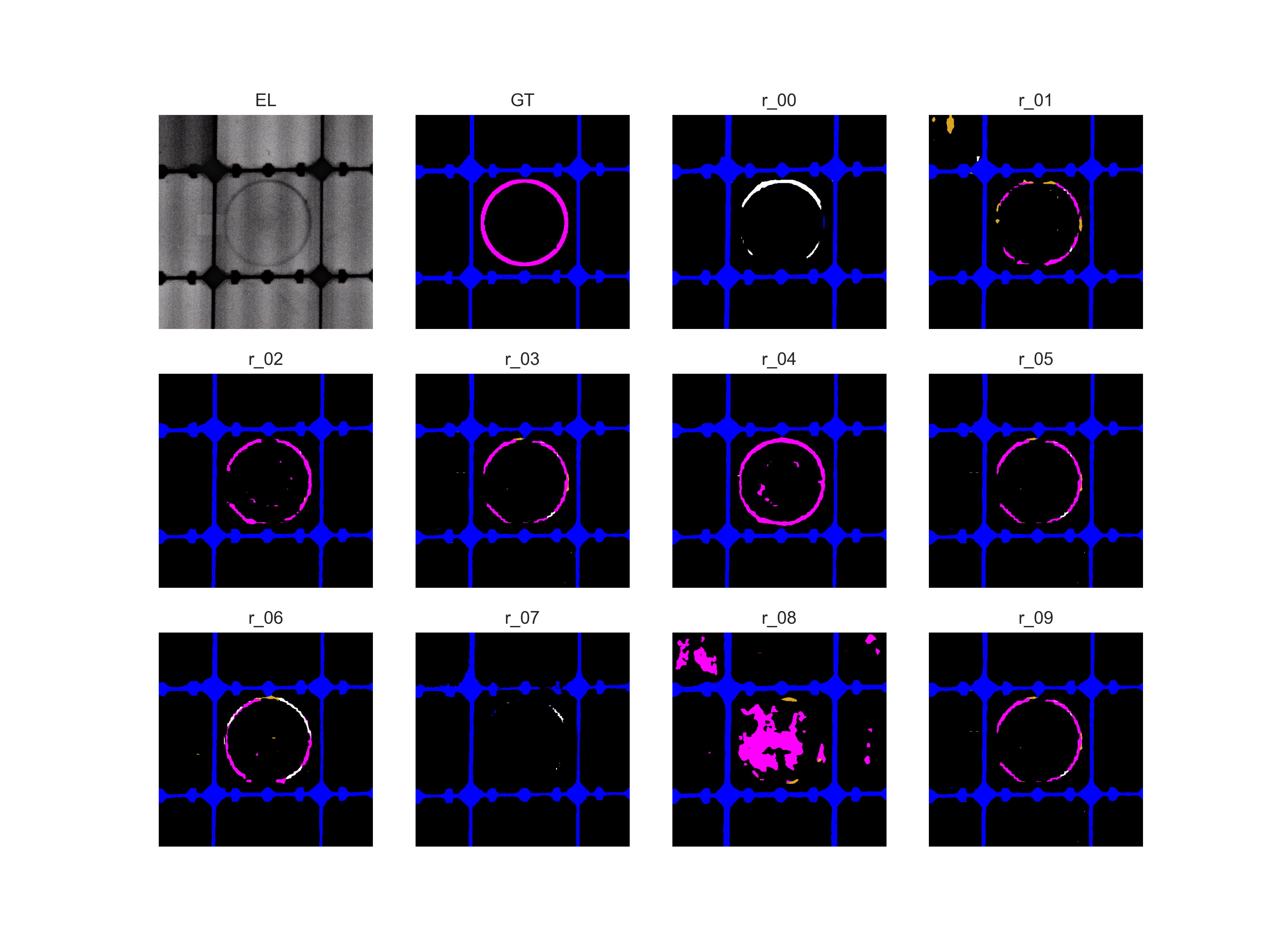}
    \caption{EL image, ground truth, and predictions for CSIR\_00067\_r3\_c2 with ring defect and dogbone spacing}
    \label{fig:csir_00067}
\end{figure}

Many of the features and defects in the dataset are underrepresented, so training and statistical analysis are challenging. However, several of the models perform well on underrepresented defects, even for defects not detected by the R0 SOTA model. For example, the `ring' defect was present in only two images from the training dataset and one image in the test dataset. The `dogbone spacing' feature (blue) was only present in 14 images. Figure \ref{fig:csir_00067} shows the EL, ground truth, and predictions from the R0 SOTA and the nine new models for the ring defect and dogbone spacing feature. While the R7 CCT model performed among the best overall for defects, the ring defect was only partially detected and incorrectly classified. Despite the small sample size, the average IoU for the ring defect was significantly higher for R4 compared to every other model ($p < 0.04$).

\subsection{OOD Analysis}
It is clear from Table \ref{tbl:ood_results} that OOD pretraining overall resulted in better performance. The previous R0 SOTA model is excluded from this table in order to standardise on models implemented and analysed in this work. It is interesting to note that this result generalises to the SCDD domain, even in the case where the in-domain unlabelled dataset for pretraining $X_{ul}$ contains a non-trivial number of examples -- approximately 30k (although still less than the OOD dataset). The results suggest that the scale of the pretraining dataset (in our experiments, ImageNet, and to a lesser degree, COCO) matters more for final performance than whether the dataset is in-distribution or OOD. We posit that the features learned when pretraining at scale result in richer, more useful embeddings and backbone networks than in-distribution pretraining on a smaller dataset.

\begin{table}[!h]
\centering
\caption{Results over all models by whether an OOD dataset was used for pretraining.}
\label{tbl:ood_results}
\begin{tabular}{|c|l|l|}
\hline
OOD   & mIoU  & wIoU  \\ \hline
\xmark & 0.505 & 0.557 \\ \hline
\cmark  & \textbf{0.530} & \textbf{0.573} \\ \hline
\end{tabular}
\end{table}

SelfSL pretraining within regimes R4 and R6 provides more evidence for the advantage of pretraining on large OOD datasets. The ImageNet-pretrained SimCLR model (R4) performed better than the EL-pretrained SimCLR model (R6). We verify this result by performing an analogous experiment on the Cityscapes dataset. We take $90\%$ of the training set for SimCLR-based SelfSL pretraining (with the same training parameters as R6), and the remaining $10\%$ for fine-tuning in a DeepLabV3 model. The results of this can be found in Table \ref{tbl:cityscapes}.

\begin{table}[!h]
\centering
\caption{Results for OOD and in-distribution pretraining on the Cityscapes and EL datasets (as per R6).}
\label{tbl:cityscapes}
\begin{tabular}{|c|c|l|l|}
\cline{2-4}
\multicolumn{1}{l|}{}      & OOD & mIoU           & wIoU           \\ \hline
\multirow{2}{*}{Cityscapes} & \xmark  & 0.189          & 0.252          \\ \cline{2-4} 
                            & \cmark & \textbf{0.243} & \textbf{0.319} \\ \hline \hline
\multirow{2}{*}{EL (R4 \& R6)}    & \xmark  & 0.516          & 0.571          \\ \cline{2-4} 
                            & \cmark & \textbf{0.534} & \textbf{0.578} \\ \hline
\end{tabular}
\end{table}

A primary reason for this performance gap when pretraining on the in-distribution dataset is likely due to an observation that SelfSL methods perform significantly better when trained on object-centric, clean datasets \citep{Purushwalkam020}, such as ImageNet. Semantic segmentation datasets typically contain multi-object images, as Cityscapes and EL indeed do. Thus, they are likely not as suitable for SelfSL pretraining. The reason for this is that contrastive self-supervised methods, such as SimCLR, learn an objective whereby the embeddings from two random views of the same image attract, and two different views repel. However, in a non-object-centric setting such as the EL or Cityscapes datasets, two random views from the same image may contain different objects/defects/features, but the network's objective would essentially be encouraging them to be the `same' (i.e. represented by the same embedding). Our results suggest that this property results in less useful representations for the downstream semantic segmentation task.

Further, the particularly good performance of pretraining on the COCO dataset in a supervised manner (R2) -- even outperforming supervised pretraining on ImageNet in many cases -- is likely due to the fact that this is pretraining on the task of semantic segmentation, rather than on image classification. This would result in a pretrained network, and resulting embeddings, more tailored towards the semantic segmentation task of SCDD.

\section{Conclusion}
\label{sec:conclusion}
The experimental results suggest that supervised training on a large OOD dataset (COCO), self-supervised pretraining on a large OOD dataset (ImageNet), and in-domain semi-supervised pretraining (CCT) all yield statistically equivalent performance for mIoU. We give preference to the COCO-pretrained model due to its comparative simplicity, which would make the deployment and maintenance of such a system significantly easier. The semi-supervised pretraining (CCT) using the relatively small domain-specific images did not improve the model performance compared to OOD pretraining, contrary to recent results in more object-centric domains. However, some DeepLabV3 models with a ResNet50 backbone achieved a new state-of-the-art for SCDD using semantic segmentation. The results also demonstrate that certain models result in a statistically significant improvement for crack detection and underrepresented classes compared to state-of-the-art based on a DeepLabV3+ model with a ResNet34 backbone. We posit that the multiview nature of various SelfSL/SemiSL techniques is ill-suited to the EL semantic segmentation task and that future work in this field may yield improved techniques to pretrain such models using large amounts of unlabelled EL data. To this end, we also contribute a new EL image dataset of $22000$ unlabelled images, for further research in developing machine learning models for semantic segmentation in the EL image domain.

\bibliographystyle{unsrtnat}
\bibliography{references}

\begin{thebibliography}{34}
\providecommand{\natexlab}[1]{#1}
\providecommand{\url}[1]{\texttt{#1}}
\expandafter\ifx\csname urlstyle\endcsname\relax
  \providecommand{\doi}[1]{doi: #1}\else
  \providecommand{\doi}{doi: \begingroup \urlstyle{rm}\Url}\fi

\bibitem[Agency(2022)]{iea2022pv}
International~Energy Agency.
\newblock Solar {PV}, {IEA}.
\newblock https://www.iea.org/reports/solar-pv, 2022.

\bibitem[Pratt et~al.(2021)Pratt, Govender, and Klein]{pratt2021}
Lawrence Pratt, Devashen Govender, and Richard Klein.
\newblock Defect detection and quantification in electroluminescence images of solar pv modules using u-net semantic segmentation.
\newblock \emph{Renewable Energy}, 2021.

\bibitem[Pratt et~al.(2023)Pratt, Mattheus, and Klein]{pratt2023}
Lawrence Pratt, Jana Mattheus, and Richard Klein.
\newblock A benchmark dataset for defect detection and classification in electroluminescence images of pv modules using semantic segmentation.
\newblock \emph{Systems and Soft Computing}, 2023.

\bibitem[Russakovsky et~al.(2015)Russakovsky, Deng, Su, Krause, Satheesh, Ma, Huang, Karpathy, Khosla, Bernstein, Berg, and Fei-Fei]{ILSVRC15}
Olga Russakovsky, Jia Deng, Hao Su, Jonathan Krause, Sanjeev Satheesh, Sean Ma, Zhiheng Huang, Andrej Karpathy, Aditya Khosla, Michael Bernstein, Alexander~C. Berg, and Li~Fei-Fei.
\newblock {ImageNet Large Scale Visual Recognition Challenge}.
\newblock \emph{International Journal of Computer Vision (IJCV)}, 2015.

\bibitem[Cordts et~al.(2016)Cordts, Omran, Ramos, Rehfeld, Enzweiler, Benenson, Franke, Roth, and Schiele]{Cordts2016Cityscapes}
Marius Cordts, Mohamed Omran, Sebastian Ramos, Timo Rehfeld, Markus Enzweiler, Rodrigo Benenson, Uwe Franke, Stefan Roth, and Bernt Schiele.
\newblock The cityscapes dataset for semantic urban scene understanding.
\newblock In \emph{Proc. of the IEEE Conference on Computer Vision and Pattern Recognition (CVPR)}, 2016.

\bibitem[Krizhevsky et~al.(2009)Krizhevsky, Hinton, et~al.]{krizhevsky2009learning}
Alex Krizhevsky, Geoffrey Hinton, et~al.
\newblock Learning multiple layers of features from tiny images.
\newblock 2009.

\bibitem[Chen et~al.(2020{\natexlab{a}})Chen, Kornblith, Norouzi, and Hinton]{simclr}
Ting Chen, Simon Kornblith, Mohammad Norouzi, and Geoffrey Hinton.
\newblock A simple framework for contrastive learning of visual representations.
\newblock In \emph{Proceedings of the 37th International Conference on Machine Learning}, 2020{\natexlab{a}}.

\bibitem[Chen et~al.(2020{\natexlab{b}})Chen, Fan, Girshick, and He]{mocov2}
Xinlei Chen, Haoqi Fan, Ross Girshick, and Kaiming He.
\newblock Improved baselines with momentum contrastive learning.
\newblock \emph{arXiv preprint arXiv:2003.04297}, 2020{\natexlab{b}}.

\bibitem[Ouali et~al.(2020)Ouali, Hudelot, and Tami]{cct}
Yassine Ouali, Celine Hudelot, and Myriam Tami.
\newblock Semi-supervised semantic segmentation with cross-consistency training.
\newblock In \emph{The IEEE/CVF Conference on Computer Vision and Pattern Recognition (CVPR)}, 2020.

\bibitem[Wang et~al.(2022)Wang, Wang, Shen, Fei, Li, Jin, Wu, Zhao, and Le]{u2pl}
Yuchao Wang, Haochen Wang, Yujun Shen, Jingjing Fei, Wei Li, Guoqiang Jin, Liwei Wu, Rui Zhao, and Xinyi Le.
\newblock Semi-supervised semantic segmentation using unreliable pseudo labels.
\newblock In \emph{Proceedings of the IEEE/CVF International Conference on Computer Vision and Pattern Recognition (CVPR)}, 2022.

\bibitem[Ericsson et~al.(2020)Ericsson, Gouk, and Hospedales]{sslanalysis}
Linus Ericsson, Henry G.~R. Gouk, and Timothy~M. Hospedales.
\newblock How well do self-supervised models transfer?
\newblock \emph{2021 IEEE/CVF Conference on Computer Vision and Pattern Recognition (CVPR)}, 2020.

\bibitem[Lin et~al.(2014)Lin, Maire, Belongie, Hays, Perona, Ramanan, Doll{\'a}r, and Zitnick]{lin2014microsoft}
Tsung-Yi Lin, Michael Maire, Serge Belongie, James Hays, Pietro Perona, Deva Ramanan, Piotr Doll{\'a}r, and C~Lawrence Zitnick.
\newblock Microsoft coco: Common objects in context.
\newblock In \emph{Computer Vision--ECCV 2014: 13th European Conference, Zurich, Switzerland, September 6-12, 2014, Proceedings, Part V 13}, 2014.

\bibitem[French et~al.()French, Karimi, and Braid]{French_Karimi_Braid_2020}
Roger~H French, Ahmad~M Karimi, and Jennifer~L Braid.
\newblock Electroluminescent (el) image dataset of pv module under step-wise damp heat exposures.
\newblock URL \url{osf.io/4qrtv}.

\bibitem[Buerhop-Lutz et~al.(2018)Buerhop-Lutz, Deitsch, Maier, Gallwitz, Berger, Doll, Hauch, Camus, and Brabec]{Buerhop2018}
Claudia Buerhop-Lutz, Sergiu Deitsch, Andreas Maier, Florian Gallwitz, Stephan Berger, Bernd Doll, Jens Hauch, Christian Camus, and Christoph~J. Brabec.
\newblock A benchmark for visual identification of defective solar cells in electroluminescence imagery.
\newblock In \emph{European PV Solar Energy Conference and Exhibition (EU PVSEC)}, 2018.

\bibitem[Deitsch et~al.(2019)Deitsch, Christlein, Berger, Buerhop-Lutz, Maier, Gallwitz, and Riess]{Deitsch2019}
Sergiu Deitsch, Vincent Christlein, Stephan Berger, Claudia Buerhop-Lutz, Andreas Maier, Florian Gallwitz, and Christian Riess.
\newblock Automatic classification of defective photovoltaic module cells in electroluminescence images.
\newblock \emph{Solar Energy}, 2019.

\bibitem[Deitsch et~al.()Deitsch, Buerhop-Lutz, Sovetkin, Steland, Maier, Gallwitz, and Riess]{Deitsch2021}
Sergiu Deitsch, Claudia Buerhop-Lutz, Evgenii Sovetkin, Ansgar Steland, Andreas Maier, Florian Gallwitz, and Christian Riess.
\newblock Segmentation of photovoltaic module cells in uncalibrated electroluminescence images.

\bibitem[pvE(2021)]{pvELad}
Pv el anomoly detection database, 2021.
\newblock URL \url{https://www.kaggle.com/competitions/pvelad/overview}.

\bibitem[Fioresi et~al.(2022)Fioresi, Colvin, Frota, Gupta, Li, Seigneur, Vyas, Oliveira, Shah, and Davis]{fioresi2022automated}
Joseph Fioresi, Dylan~J. Colvin, Rafaela Frota, Rohit Gupta, Mengjie Li, Hubert~P. Seigneur, Shruti Vyas, Sofia Oliveira, Mubarak Shah, and Kristopher~O. Davis.
\newblock Automated defect detection and localization in photovoltaic cells using semantic segmentation of electroluminescence images.
\newblock \emph{IEEE Journal of Photovoltaics}, 2022.

\bibitem[Tsai et~al.(2012)Tsai, Wu, and Li]{tsai2012defect}
Du-Ming Tsai, Shih-Chieh Wu, and Wei-Chen Li.
\newblock Defect detection of solar cells in electroluminescence images using fourier image reconstruction.
\newblock \emph{Solar Energy Materials and Solar Cells}, 2012.

\bibitem[Spataru et~al.(2016)Spataru, Hacke, and Sera]{spataru2016automatic}
Sergiu Spataru, Peter Hacke, and Dezso Sera.
\newblock Automatic detection and evaluation of solar cell micro-cracks in electroluminescence images using matched filters.
\newblock In \emph{2016 IEEE 43rd Photovoltaic Specialists Conference (PVSC)}, 2016.

\bibitem[Chen et~al.(2019)Chen, Zhao, Han, and Liu]{chen2019accurate}
Haiyong Chen, Huifang Zhao, Da~Han, and Kun Liu.
\newblock Accurate and robust crack detection using steerable evidence filtering in electroluminescence images of solar cells.
\newblock \emph{Optics and Lasers in Engineering}, 2019.

\bibitem[Rahman and Chen(2020)]{rahman2020defects}
Muhammad Rameez~Ur Rahman and Haiyong Chen.
\newblock Defects inspection in polycrystalline solar cells electroluminescence images using deep learning.
\newblock \emph{IEEE Access}, 2020.

\bibitem[Sohn(2016)]{infonce}
Kihyuk Sohn.
\newblock Improved deep metric learning with multi-class n-pair loss objective.
\newblock In \emph{Advances in Neural Information Processing Systems}, 2016.

\bibitem[He et~al.(2020)He, Fan, Wu, Xie, and Girshick]{moco}
Kaiming He, Haoqi Fan, Yuxin Wu, Saining Xie, and Ross Girshick.
\newblock Momentum contrast for unsupervised visual representation learning.
\newblock In \emph{2020 IEEE/CVF Conference on Computer Vision and Pattern Recognition (CVPR)}, 2020.

\bibitem[Zhang et~al.(2023)Zhang, Zheng, and Gu]{ssl-for-medical}
Chuyan Zhang, Hao Zheng, and Yun Gu.
\newblock Dive into the details of self-supervised learning for medical image analysis.
\newblock \emph{Medical Image Analysis}, 89:\penalty0 102879, 2023.
\newblock ISSN 1361-8415.
\newblock \doi{https://doi.org/10.1016/j.media.2023.102879}.

\bibitem[Torpey et~al.(2024)Torpey, Parkin, Alter, Klein, and James]{phd-jei-midas}
David Torpey, Max Parkin, Jonah Alter, Richard Klein, and Steven James.
\newblock {MiDaS: a large-scale Minecraft dataset for non-natural image benchmarking}.
\newblock \emph{Journal of Electronic Imaging}, 33\penalty0 (1):\penalty0 013035, 2024.
\newblock \doi{10.1117/1.JEI.33.1.013035}.

\bibitem[Lee(2013)]{Pseudolabel}
Dong-Hyun Lee.
\newblock Pseudo-label : The simple and efficient semi-supervised learning method for deep neural networks.
\newblock \emph{ICML 2013 Workshop : Challenges in Representation Learning (WREPL)}, 2013.

\bibitem[He et~al.(2016)He, Zhang, Ren, and Sun]{resnet}
Kaiming He, Xiangyu Zhang, Shaoqing Ren, and Jian Sun.
\newblock Deep residual learning for image recognition.
\newblock In \emph{Proceedings of 2016 IEEE Conference on Computer Vision and Pattern Recognition}, 2016.

\bibitem[maintainers and contributors(2016)]{torchvision}
TorchVision maintainers and contributors.
\newblock Torchvision: Pytorch's computer vision library.
\newblock https://github.com/pytorch/vision, 2016.

\bibitem[Goyal et~al.(2021)Goyal, Duval, Reizenstein, Leavitt, Xu, Lefaudeux, Singh, Reis, Caron, Bojanowski, Joulin, and Misra]{vissl}
Priya Goyal, Quentin Duval, Jeremy Reizenstein, Matthew Leavitt, Min Xu, Benjamin Lefaudeux, Mannat Singh, Vinicius Reis, Mathilde Caron, Piotr Bojanowski, Armand Joulin, and Ishan Misra.
\newblock Vissl.
\newblock https://github.com/facebookresearch/vissl, 2021.

\bibitem[You et~al.(2017)You, Gitman, and Ginsburg]{lars}
Yang You, Igor Gitman, and Boris Ginsburg.
\newblock Large batch training of convolutional networks, 2017.

\bibitem[Seabold and Perktold(2010{\natexlab{a}})]{seabold2010statsmodels}
Skipper Seabold and Josef Perktold.
\newblock statsmodels: Econometric and statistical modeling with python.
\newblock In \emph{9th Python in Science Conference}, 2010{\natexlab{a}}.

\bibitem[Seabold and Perktold(2010{\natexlab{b}})]{multicomp}
Skipper Seabold and Josef Perktold.
\newblock statsmodels: Econometric and statistical modeling with python.
\newblock In \emph{9th Python in Science Conference}, 2010{\natexlab{b}}.

\bibitem[Purushwalkam and Gupta(2020)]{Purushwalkam020}
Senthil Purushwalkam and Abhinav Gupta.
\newblock Demystifying contrastive self-supervised learning: Invariances, augmentations and dataset biases.
\newblock In \emph{Advances in Neural Information Processing Systems 33: Annual Conference on Neural Information Processing Systems 2020, NeurIPS 2020, December 6-12, 2020, virtual}, 2020.

\end{thebibliography}

\end{document}